\pdfoutput=1

\documentclass[11pt]{article}

\usepackage{emnlp2021}

\usepackage{times}
\usepackage{latexsym}

\usepackage[T1]{fontenc}

\usepackage[utf8]{inputenc}

\usepackage{microtype}

\usepackage{soul}
\usepackage{url}
\usepackage[utf8]{inputenc}
\usepackage{graphicx}
\usepackage{amsmath}
\usepackage{booktabs}
\usepackage{algorithm}
\usepackage{algorithmic}
\usepackage{epstopdf}
\usepackage{multirow}
\usepackage{makecell}
\usepackage{graphicx}
\usepackage{enumitem}
\usepackage{epstopdf}
\usepackage{appendix}
\usepackage{multirow}
\usepackage{makecell}
\usepackage{bm}
\usepackage{soul,color}
\usepackage{amssymb}

\title{Hyperbolic Geometry is Not Necessary: Lightweight Euclidean-Based Models for Low-Dimensional Knowledge Graph Embeddings}

\author{
Kai Wang$^{1,2,3}$\thanks{~~The Corresponding Author}, Yu Liu$^{1,2}$, Dan Lin$^{1,2}$, Quan Z. Sheng$^3$\\
$^1$School of Software Technology, Dalian University of Technology, Dalian, 116620, China\\
$^2$Key Laboratory for Ubiquitous Network and Service Software of Liaoning Province, China\\
{\tt kai\_wang@mail.dlut.edu.cn, yuliu@dlut.edu.cn, lindan0823@mail.dlut.edu.cn}\\
$^3$Department of Computing, Macquarie University, Sydney, NSW 2109, Australia\\
{\tt michael.sheng@mq.edu.au}
}

\begin{document}
\maketitle
\begin{abstract}
Recent knowledge graph embedding (KGE) models based on hyperbolic geometry have shown great potential in a low-dimensional embedding space. However, the necessity of hyperbolic space in KGE is still questionable, because the calculation based on hyperbolic geometry is much more complicated than Euclidean operations. In this paper, based on the state-of-the-art hyperbolic-based model RotH, we develop two lightweight Euclidean-based models, called {\em RotL} and {\em Rot2L}. The RotL model simplifies the hyperbolic operations while keeping the flexible normalization effect. Utilizing a novel two-layer stacked transformation 
and based on RotL, the Rot2L model obtains an improved representation capability,
yet costs fewer parameters and calculations than RotH. The experiments on link prediction show that Rot2L achieves the state-of-the-art performance on two widely-used datasets in low-dimensional knowledge graph embeddings. Furthermore, RotL achieves similar performance as RotH but only requires half of the training time.
\end{abstract}

\section{Introduction}
To represent entities and relations of knowledge graphs (KGs) in the semantic vector space, researchers have proposed various knowledge graph embedding (KGE) models, which have shown great potential in knowledge graph completion and knowledge-driven applications \cite{2017Survey,ACL20KGE1}.
To achieve higher prediction accuracy, recent KGE models usually use high-dimensional embedding vectors up to 200 or even 500 dimensions \cite{RotatE, QuatE}. 
However, when facing large-scale KGs with millions 
of entities, high embedding dimensions would require prohibitive training costs and storage space \cite{KGCompress,ACL20KGE2}.
It hinders the practical application of KGE models, especially in mobile smart devices.

\begin{figure}[!bt]
\centering
\includegraphics[width=0.45\textwidth]{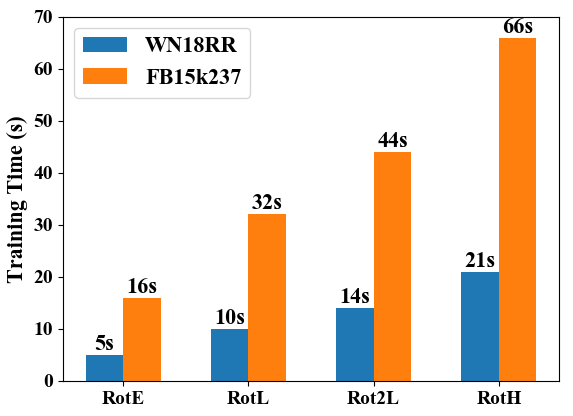}
\caption{The training time per epoch of different 32-dimensional models on two datasets. All results are measured under the same parameter settings (500 batch size and 500 negative samples). RotL and Rot2L are the models proposed in this paper.}
\label{fig:1}
\vspace{-2mm}
\end{figure}

Recently, low-dimensional KGE models based on hyperbolic vector space have drawn some attention~\cite{EMNLP20HBKGE}.
The work 
of the first 
such model, MuRP, indicates that hyperbolic embeddings can capture hierarchical patterns in KGs and generate high-fidelity and parsimonious representations \cite{MuRP}.
To capture logical patterns in KGs, Chami et al. propose a series of hyperbolic KGE models, including RotH, RefH, and AttH \cite{GoogleAttH}.
Similar to the typical TransE model \cite{TransE} that treats the relation as a translation operation between the head and tail entity vectors, the state-of-the-art RotH model adjusts the head vector by the rotation and translation transformations to approach the tail vector in the hyperbolic space.

Although the above hyperbolic-based models outperform previous Euclidean-based models in low-dimensional condition, the necessity of hyperbolic space in this task is still questionable.
Comparing a hyperbolic model with its Euclidean-based variant,
it is uncertain which parts of the modification will be vital.
Besides, despite theoretical support, the M{\"o}bius matrix-vector multiplication and M{\"o}bius addition operations in hyperbolic embeddings are far more complicated than Euclidean multiplication and addition. 
As shown in Fig. \ref{fig:1}, RotH requires threefold more training time than its Euclidean-based variant RotE on two datasets.
Especially on large-scale knowledge graphs, the additional calculating cost caused by the complicated hyperbolic operations would make the problem much severer.

Facing these problems, we 
analyze the effective components in the hyperbolic KGE models, and propose two lightweight ``RotH-like'' models, RotL and Rot2L, for low-dimensional knowledge graph embeddings. 
Without the hyperbolic geometry, RotL eliminates the M{\"o}bius matrix-vector multiplication and designs a new flexible addition operation to replace the M{\"o}bius addition.
To further improve the RotL's representation capability, the 
Rot2L model utilizes two stacked rotation-translation transformations in the Euclidean space. 
Benefiting from a specific parameterization strategy, Rot2L requires fewer parameters and calculations than RotH.

We conduct extensive experiments on two widely-used datasets.
The results show that RotL outperforms existing Euclidean-based models in the 32-dimensional condition and only requires half of the training time of RotH. Rot2L obtains the state-of-the-art performance on the two datasets and outperforms RotH in both prediction accuracy and training speed. 
According to ablation experiments, we prove the effectiveness of the flexible addition and the other significant modules in Rot2L. 
We also verify our models in different embedding dimensions and 
analyze the performance difference between RotH and our models in a relation-specific experiment.

The rest of the paper is organized as follows. We discuss the background and definitions in Sec. \ref{sec:2}.
Sec. \ref{sec:3} introduces the 
technical details of RotL and Rot2L models. Sec. \ref{sec:4} reports the experimental studies and Sec. \ref{sec:5} further discusses several experimental investigations. The related work is reviewed 
in Sec. \ref{sec:6}. Finally, we offer some concluding remarks in Sec. \ref{sec:7}.

\section{Background}
\label{sec:2}
In this section, we briefly describe the preliminaries 
related to this work. 

\subsection{Knowledge Graph Embeddings}
In a knowledge graph $\mathcal{G}=(E,R,T)$, $E$ and $R$ denote the set of entities and relations, and $T$ is the collection of factual triples $(h,r,t)$ where the head and tail entities $h, t \in E$ and the relation $r \in R$.
$N_e$ and $N_r$ refer to the number of entities and relations, respectively.

Knowledge graph embeddings aim to represent each entity $e$ and each relation $r$ as $d$-dimensional continuous vectors.
A KGE model is evaluated by the link prediction task, which 
aims to find $e_t \in E$ given an e-r query $q = (e, r)$, such that the triple $(e, r, e_t)$ or $(e_t, r, e)$ should belong to the knowledge graph $\mathcal{G}$.
Generally, a scoring function $F(h, r, t)$ is designed to 
measure each candidate triple.
Take the distance-based scoring function $F(h, r, t)=D(Q(h,r), t)$ as an example, it 
involves 
two operations:
1) {\em Transformation function} $Q(h,r)$ transforms the head vector $\bm{h}$ using the relation vector $\bm{r}$;
2) {\em Distance function} $D(\bm{q},t)$ measures the distance between the tail vector $\bm{t}$ and the transformed head vector $\bm{q}=Q(h,r)$.

\subsection{Hyperbolic Geometry}
Recently, researchers start to work on effective low-dimensional models in the KGE domain \cite{KGCompress,OurWWW, OurIJCAI}.
Multiple hyperbolic KGE models, such as MuRP, RotH, RefH and AttH, 
have achieved good performance in low-dimensional condition \cite{MuRP,GoogleAttH}.
These models employ a hyperbolic geometry model, the $d$-dimensional Poincar\'{e} ball \cite{HBMath2} with negative curvature $\textnormal{-}c$ $(c \textgreater 0)$: $\mathbb{B}_c^d=\{\mathbf{x} \in \mathbb{R}^{d}:\|\mathbf{x}\|^{2}<\frac{1}{c}\}$, where $\| \cdot \|$ denotes the $L_2$ norm.

The hyperbolic space is one of the three kinds of isotropic spaces, and the relevant theoretical research has been carried out for decades \cite{HBMath2}.
To achieve the vector transformation in the hyperbolic space, the M{\"o}bius addition $\oplus_c$ and M{\"o}bius matrix-vector multiplication $\otimes_c$ are utilized.
M{\"o}bius addition \cite{HBMath1} is proposed to approximate Euclidean addition in the hyperbolic space: 
\vspace{-0.2cm}
\begin{equation}
\footnotesize
\mathbf{x} \oplus_{c} \mathbf{y}=\frac{\left(1+2 c\langle\mathbf{x}, \mathbf{y}\rangle+c\|\mathbf{y}\|^{2}\right) \mathbf{x}+\left(1-c\|\mathbf{x}\|^{2}\right) \mathbf{y}}{1+2 c\langle\mathbf{x}, \mathbf{y}\rangle+c^{2}\|\mathbf{x}\|^{2}\|\mathbf{y}\|^{2}}
\end{equation}
where $\langle \cdot \rangle$ is the Euclidean inner product. It is clear that M{\"o}bius addition requires much more calculations than an ordinary addition.

M{\"o}bius matrix-vector multiplication \cite{HNN} is also more complicated than Euclidean multiplication. Before computing matrix multiplication with $M \in \mathbb{R}^{d \times k}$, the vector $x \in \mathbb{B}_c^d $ is projected onto the tangent space at $\mathbf{0} \in \mathbb{B}_c^d$ with the \textit{logarithmic map} $log_0^c(x)$. Then the output of multiplication is projected back to $\mathbb{B}_c^d$ via the \textit{exponential map} $exp_0^c(x)$, i.e.,
\begin{equation}
\footnotesize
M \otimes_c \bm{x} = exp_0^c(Mlog_0^c(\bm{x}))
\vspace{-5mm}
\end{equation}
\begin{equation}
\footnotesize
\exp _{0}^{c}(\mathbf{x})=\tanh (\sqrt{c}\|\mathbf{x}\|) \frac{\mathbf{x}}{\sqrt{c}\|\mathbf{x}\|}
\vspace{-3mm}
\end{equation}
\begin{equation}
\footnotesize
\log _{0}^{c}(\mathbf{x})=\operatorname{arctanh}(\sqrt{c}\|\mathbf{x}\|) \frac{\mathbf{x}}{\sqrt{c}\|\mathbf{x}\|}
\vspace{2mm}
\end{equation}

\subsection{The RotH Model}
We briefly review RotH \cite{GoogleAttH}, the state-of-the-art model in the low-dimensional KGE.
According to the official PyTorch implementation\footnote{\url{https://github.com/HazyResearch/KGEmb}}, the scoring function of RotH employs a ``translation-rotation-translation'' transformation and utilizes a hyperbolic distance as the distance function $D$.

Specifically, let $\bm{e}^H \in \mathbb{B}_c^d$ denote entity hyperbolic embeddings of entity $e$.
For one relation $r$, two hyperbolic relation vectors $\bm{r}^H, \bm{r'}^H \in \mathbb{B}_c^d$ are defined for two translation operations.
Using a $d$-dimensional vector $\bm{\hat{r}}$, RotH parameterizes a {\em Givens Rotation} operation with a block-diagonal matrix of the form:
\begin{equation}
\footnotesize
Rot(\bm{\hat{r}}) = diag(G(\hat{r}_1, \hat{r}_2), \dots, G(\hat{r}_{d-1}, \hat{r}_d))
\vspace{-7mm}
\end{equation}
\begin{equation}
\footnotesize
where \ \  G(\hat{r}_i, \hat{r}_j) := \left[\begin{array}{ll}
\hat{r}_i,& \textnormal{-}\hat{r}_j \\
\hat{r}_j,& \hat{r}_i
\end{array}\right].
\end{equation}

Then, 
for a triple $(h, r, t)$, the scoring function $F_H$ of RotH is defined as:
\begin{equation}
\footnotesize
Q_H^c(h, r) = Rot(\bm{\hat{r}}) \otimes_c (\bm{h}^H \oplus_c \bm{r}^H) \oplus_c \bm{r'}^H 
\vspace{-5mm}
\end{equation}
\begin{equation}
\footnotesize
D_H^c(\bm{q}, t) = -\frac{2}{\sqrt{c}}arctanh(\sqrt{c}\|-\bm{q} \oplus_c \bm{t}^H\|)^2 \label{eq0}
\vspace{-3mm}
\end{equation}
\begin{equation}
\footnotesize
F_H(h,r,t) = D_H^{c_r}(Q_H^{c_r}(h,r),t) + b_{h} + b_{t},
\label{eq1}
\end{equation}
where $c_r \textgreater 0$ is the relation-specific curvature parameter, and $b_e (e\in E)$ are entity biases which act as margins in the scoring function \cite{MuRP,GoogleAttH}.

The other hyperbolic models can be regarded as RotH's variants using different relation transformations. In addition, RotE is 
a
Euclidean-based RotH variant, and its scoring function is defined as:
\begin{equation}
\footnotesize
F_E(h, r, t) = -\|(Rot(\bm{\hat{r}})\bm{h} + \bm{r})-\bm{t}\|^2 + b_{h} + b_{t},
\label{eq2}
\end{equation}
where $\bm{h}, \bm{r}, \bm{t} \in \mathbb{R}^d$. Without complex hyperbolic calculations, $F_E$ can be computed in linear time of the embedding dimensions.

\section{The Methodology}
\label{sec:3}
The goal of this work is to design a high-efficiency low-dimensional KGE model by extracting the effective components in the RotH model and eliminating the redundancy.

We find that RotH performs noticeably well because of two reasons. 
The first reason is {\em rotation-translation transformation}. As proved in previous research \cite{RotatE,GoogleAttH}, this specific transforming operation can infer different relation patterns in the KG.
The second reason is {\em flexible normalization}. All entity vectors in the hyperbolic space satisfy $\|\mathbf{e}\|^{2}<\frac{1}{c}$ before and after transformation, while the curvature $c$ is relation-specific and self-adaptive. 
As the representation capability of a low-dimensional vector space is limited, the effect of flexible normalization would be more obvious. It explains why RotH can outperform its Euclidean-based variant RotE in low-dimensional KGE tasks.

\begin{figure*}[!bt]
\centering
\includegraphics[width=1.0\textwidth]{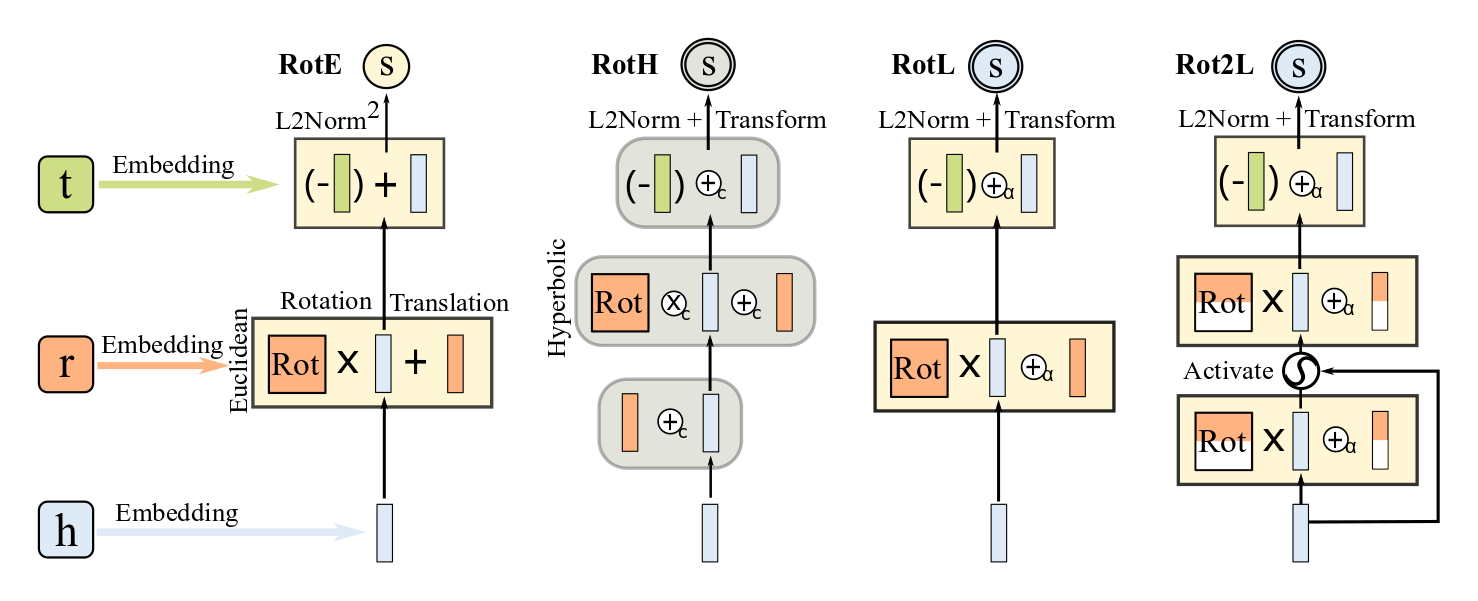}
\caption{The architectures of four models, including the previous RotE and RotH, and the proposed RotL and Rot2L in this paper. 
The rectangle box denotes 
a
Euclidean-based operation, while the rounded rectangle box denotes a hyperbolic-based one. The inside rectangles denotes the embedding vectors or matrices, in which the relation-specific ones are in orange. $\bigotimes_c, \bigoplus_c, \bigoplus_\alpha$ refer to M{\"o}bius multiplication, M{\"o}bius addition and Flexible addition, respectively.}
\label{fig:2}
\vspace{-2mm}
\end{figure*}

In this section, 
we first propose a lightweight model, called RotL, which remains the flexible normalization of RotH and simplifies the complex hyperbolic operations. The details of RotL will be described in Sec. \ref{sec:3.1}. 
We further design the Rot2L model using two stacked rotation-translation transformations. 
Rot2L employs a novel parameterization strategy that can save half of parameters in the two-layer architecture, which is detailed in Sec. \ref{sec:3.2}.
The architectures of the four models mentioned above are illustrated in Fig. \ref{fig:2}.

\subsection{The RotL Model and Flexible Addition}
\label{sec:3.1}
The RotL model aims to achieve similar performance to RotH and minimize its computational complexity close to that of RotE.
Comparing the scoring functions of RotH and RotE in Eq. \ref{eq1} and \ref{eq2}, it is clear that the additional calculations of RotH are centered on M{\"o}bius addition and M{\"o}bius matrix-vector multiplication.

Therefore, we first eliminate the hyperbolic embeddings in RotL and initialize the entity vector $\bm{e}$ and the two relation vectors for rotation and translation in the Euclidean space, such that the relation transformation can be calculated 
using Euclidean addition and multiplication directly. 

To achieve the flexible normalization, we propose Flexible Addition $\oplus_\alpha$, a simplified form of M{\"o}bius Addition, i.e.,
\begin{equation}
\footnotesize
\mathbf{x} \oplus_\alpha \mathbf{y} = \frac{\alpha(\mathbf{x}+\mathbf{y})}{1+\langle\mathbf{x}, \mathbf{y}\rangle},
\end{equation}
where $\alpha$ is a relation-specific scaling parameter and 
with a default value of 1.
The Flexible Addition provides a self-adaptive normalization to $(x+y)$, and 
has lower computational complexity than M{\"o}bius addition.
Counting the operation times of $d$-dimensional vector operations,
the former requires three additions and two multiplications, while the latter needs 
nine and 12 operations, respectively.
We further 
discuss 
the connection between the two operations through Theorem 1.

\vspace{0.2cm}
\noindent\textbf{Theorem 1.} Given that $c=\alpha=1$, the M{\"o}bius Addition $\oplus_c$ and Flexible Addition $\oplus_{\alpha}$ satisfy that $\mathbf{x} \oplus_c \mathbf{x} \equiv \mathbf{x} \oplus_\alpha \mathbf{x}$.

\vspace{0.2cm}
\noindent\emph{Proof.} With $c=\alpha=1$ and two vectors $\mathbf{x},\mathbf{x'} \in \mathcal{R}^d$, satisfying $\mathbf{x} = \mathbf{x'}$, 
\begin{equation}
\footnotesize
\begin{aligned}
&\mathbf{x} \oplus_c \mathbf{x'} = \frac{\left(1+2 \langle\mathbf{x}, \mathbf{x'}\rangle+\|\mathbf{x'}\|^{2}\right) \mathbf{x}+\left(1-\|\mathbf{x}\|^{2}\right) \mathbf{x'}}{1+2\langle\mathbf{x}, \mathbf{x'}\rangle+\|\mathbf{x}\|^{2}\|\mathbf{x'}\|^{2}}\\
&= \frac{\left(1+\langle\mathbf{x}, \mathbf{x'}\rangle + \langle\mathbf{x'}, \mathbf{x}\rangle+\|\mathbf{x'}\|^{2}\right) \mathbf{x}+\left(1-\|\mathbf{x}\|^{2}\right) \mathbf{x'}}{1+\langle\mathbf{x}, \mathbf{x'}\rangle+ \langle\mathbf{x'}, \mathbf{x}\rangle+\|\mathbf{x}\|^{2}\|\mathbf{x'}\|^{2}}\\
&= \frac{(1+\langle\mathbf{x'}, \mathbf{x}\rangle)(\mathbf{x}+\mathbf{x'})}{(1+\langle\mathbf{x}, \mathbf{x'}\rangle)(1+\langle\mathbf{x'}, \mathbf{x}\rangle)} 
=\frac{\mathbf{x}+\mathbf{x'}}{1+\langle\mathbf{x}, \mathbf{x'}\rangle} = \mathbf{x} \oplus_\alpha \mathbf{x'}
\end{aligned}
\end{equation}

We 
emphasize
that Theorem 1 indicates the equivalence of the two operations in a special condition.
In our models, the proposed Flexible Addition is not equal to the M{\"o}bius Addition. It imitates the flexible normalization of the latter and eliminates the Hyperbolic space assumption.

We then define the transformation function of RotL as $Q_L^\alpha(h, r) = Rot(\bm{\hat r})\bm{h} \oplus_\alpha \bm{r'}$, which can be regarded as a RotE transformation using the flexible addition.
To fit this novel operation, we further modify the distance function of RotH in Eq. \ref{eq0} by designing a simpler non-linear mapping operation.
The distance function and scoring function of RotL are defined as follows:
\vspace{-1mm}
\begin{equation}
\footnotesize
D_L^\alpha(\bm{q}, t) = -\varphi(\|-\bm{q} \oplus_\alpha \bm{t}\|)
\label{eq3.1}
\vspace{-8mm}
\end{equation}

\begin{equation}
\footnotesize
F_L(h,r,t) = D_L^{\alpha'_r}(Q_L^{\alpha_r}(h,r),t) + b_{h} + b_{t},
\label{eq3}
\end{equation}
where $\alpha_r$ and $\alpha'_r$ are two different scaling parameters, and $\varphi(x) = x e^x$ is empirically discovered to replace the arctanh function in RotH with less complexity.

Comparing Eq. \ref{eq1} and \ref{eq3}, it is clear that the hyperbolic calculations are completely eliminated in the RotL model. Thus, RotL can reduce the computation complexity 
of RotH and save half of the training time as shown in Fig. \ref{fig:1}.

\subsection{The Rot2L Model and Stacked Transformation}
\label{sec:3.2}
Although the lightweight RotL maintains the flexible normalization effect, its performance is limited by the original transformation function of RotH.
In this section, we 
describe a novel Rot2L model utilizing two stacked translation-rotation transformations.

According to the theory of affine transformation \cite{AffineBook}, the two transformations can be replaced by a single one.
Therefore, inspired by neural networks, we design a two-layer architecture with an activate function in the middle, as shown in Fig. \ref{fig:2}.
The transformation function $Q_{2L}(h,r)$ in Rot2L is defined as:
\begin{equation}
\footnotesize
\varrho(\bm{h},\bm{q}) = tanh(\bm{q}) + \gamma \bm{h} 
\label{eq5}
\vspace{-5mm}
\end{equation}
\begin{equation}
\footnotesize
Q_{2L}^{\alpha1,\alpha2}(h, r) = Q_L^{\alpha1}(\varrho(\bm{h},Q_L^{\alpha2}(h, r)),r),
\end{equation}
where $\gamma$ is a hyper-parameter that balances the two parts. $Q_L^{\alpha1}$ and $Q_L^{\alpha2}$ represent two transformation layers, which are the same as the transformation function in RotL.

In the Rot2L model, the two layers need different parameters.
This would double the amount of relation parameters 
because each layer requires two $N_r \times d$ embedding matrices to represent the translation vectors and rotation matrices for all relations. 
To reduce relation parameters, 
Rot2L 
employs a novel parameterization strategy, which shares partial parameters among different relations.

Specifically, we utilize 
one embedding 
matrix
$\bm{M} \in \mathbb{R}^{N_r \times d}$ and a $d$-dimensional learnable vector $\bm{f}\in \mathbb{R}^d$ for each rotation-translation transformation layer.
Such that half of the parameters are shared in different relations by replacing another embedding matrix to the vector $\bm{f}$.
Given the vector $\bm{r} = \bm{M}[r]$ for the relation $r$, the corresponding translation vector and rotation matrix are constructed as:
\begin{equation}
\footnotesize
\bm{r'} = [r_1, f_1, r_2, f_2, \dots, r_\frac{d}{2}, f_\frac{d}{2}],
\vspace{-6mm}
\end{equation} 
\begin{equation}
\footnotesize
Rot(\hat r) = diag(G(r_{\frac{d}{2}+1}, f_{\frac{d}{2}+1}), \dots, G(r_d, f_d)).
\end{equation}

Finally, the scoring function of Rot2L contains the transformation function $Q_{2L}(h,r)$ and the same distance function as RotL, which is defined as:
\begin{equation}
\footnotesize
F_L(h,r,t) = D_L^{\alpha'_r}(Q_{2L}^{\alpha1_r,\alpha2_r}(h, r),t) + b_{h} + b_{t}.
\end{equation}

Note that, it 
might be feasible to employ more transformation layers in Rot2L like deep neural networks. There are two reasons that we do not utilize more than two layers.
First, more layers require more parameters, which goes against our original intention of being lightweight. Second, we find the vector values are gradually magnified when getting through multiple layers. Using three layers in the 
Rot2L model 
already 
suffers performance decrease. Exploring a deeper model with more effective regularization 
will be our future work.

\section{Experiments}
\label{sec:4}
\subsection{Experimental Setup}

\textbf{Datasets}. Our experimental studies are conducted on two widely-used datasets.
WN18RR \cite{WNDS} is a subset of the English lexical database WordNet \cite{WordNet}, while
FB15k237 \cite{Toutanova2015} is extracted from Freebase including knowledge facts 
on movies, actors, awards, and sports. 
Inverse relations are removed from the two datasets, as many test triples can be obtained simply by inverting triples in the training set.
The statistics of the datasets are given in Table \ref{tab:1} and 
``Train'', ``Valid'', ``Test'' refer to the amount of triples in training, validation, and test sets.

\begin{table}[!htb]
\caption{Statistics of the datasets.}
\centering
\label{tab:1}
\resizebox{\linewidth}{!}{
\begin{tabular}{c|rrrrr}
\hline
\textbf{Dataset} & \textbf{$N_r$} & \textbf{$N_e$} & \textbf{\#Train} & \textbf{\#Valid} & \textbf{\#Test} \\
\hline
FB15k237 & $237$ & $14,541$ & $272,115$ & $17,535$ & $20,466$ \\
WN18RR & $11$ & $40,943$ & $86,845$ & $3,034$ & $3,134$ \\
\hline
\end{tabular} 
}
\end{table}

\begin{table*}[!tb]
\caption{The link prediction results on the WN18RR and FB15k237 datasets. The best scores of 32-dimensional models are in \textbf{Bold}.}
\small
\centering
\label{tab:2}
\begin{tabular}{ll|ccc|ccc}
\hline
\multirow{2}*{\textbf{Type}} & \multirow{2}*{\textbf{Methods}} & \multicolumn{3}{c|}{\textbf{FB15K237}} & \multicolumn{3}{c}{\textbf{WN18RR}}\\
~ & ~ & \textbf{MRR} & \textbf{Hits@10} & \textbf{Hits@1} & \textbf{MRR} & \textbf{Hits@10} & \textbf{Hits@1}\\
\hline
\multirow{6}*{\makecell[l]{Euclidean-based \\ Models}} & RotatE & 0.290 & 0.458 & 0.208 & 0.387 & 0.491 & 0.330\\
~ & TuckER & 0.306 & 0.475 & 0.223 & 0.428 & 0.474 & 0.401\\
~ & MuRE & 0.313 & 0.489 & 0.226 & 0.458 & 0.525 & 0.421 \\
~ & RefE & 0.302 & 0.474 & 0.216 & 0.455 & 0.521 & 0.419 \\
~ & RotE & 0.307 & 0.482 & 0.220 & 0.463 & 0.529 & 0.426 \\
~ & AttE & 0.311 & 0.488 & 0.223 & 0.456 & 0.526 & 0.419\\
\hline 
\multirow{4}*{\makecell[l]{Hyperbolic-based \\ Models}} & MuRP & 0.323 & 0.501 & 0.235 & 0.465 & 0.544 & 0.420 \\
~ & RefH & 0.312 & 0.489 & 0.224 & 0.447 & 0.518 & 0.408 \\
~ & RotH & 0.314 & 0.497 & 0.223 & 0.472 & 0.553 & 0.428 \\
~ & AttH & 0.324 & 0.501 & 0.236 & 0.466 & 0.551 & 0.419 \\
\hline 
\multirow{2}*{\makecell[l]{Our Models}} & \textbf{RotL} & 0.320  & 0.500  & 0.229  & 0.469  & 0.550  & 0.426\\
~ & \textbf{Rot2L} & \textbf{0.326}  & \textbf{0.503}  & \textbf{0.237}  & \textbf{0.475}  & \textbf{0.554}  & \textbf{0.434}\\
\hline
\end{tabular}
\end{table*}

\vspace{2mm}
\noindent\textbf{Implementation Details}. 
Following the previous work, we utilize a binary cross-entropy loss, which is defined as:
\begin{equation}
\hspace{-0.5mm}
\footnotesize
L = -log\sigma(F(h,r,t))-\sum_{i=0}^k log(1-\sigma(F(h'_i,r,t'_i))),
\end{equation}
where $\sigma(\cdot)$ refers to the Sigmoid function, and ${(h'_i,r,t'_i)}$ refers to the negative samples after deleting training triples.
All experiments are performed on NVIDIA GeForce GTX1080Ti GPUs, and implemented in Python using the PyTorch framework. 

\vspace{2mm}
\noindent\textbf{Hyperparameter Settings.} 
According to the low dimensional condition, we train our model setting the embedding dimensions in $\{8, 16, 32, 64\}$, with the Adam optimizer for the WN18RR dataset and Adagrad optimizer for FB15k237.
We select the hyper-parameters of our model via grid search according to the metrics on the validation set. 
Specifically, we empirically select the learning rate among $\{0.0005, 0.005, 0.05\}$, the batch size among $\{100,200,500\}$, the amount of negative samples $k$ among $\{50,200,500\}$,
and the balance hyper-parameter $\gamma$ among $\{0.1, 0.3, 0.5, 1.0\}$.

\vspace{2mm}
\noindent\textbf{Evaluation Metrics.}
For the link prediction experiments, we adopt three evaluation metrics:
1) MRR, the average inverse rank of the test triples, 
2) Hits@10, the proportion of correct entities ranked in top 10, and
3) Hits@1, the proportion of correct entities ranked first.
Higher MRR, Hits@10, and Hits@1 mean better performance.
Following the previous works, we process the output sequence in the filter mode.

\subsection{Link Prediction Task}
We evaluate our models in the link prediction task on the two datasets.
The experimental results are shown in Table \ref{tab:2}.
We select ten compared models in two types, in which the first 
six
models are Euclidean-based, and the others utilize hyperbolic embeddings.
Following the setting of Chami et al. \cite{GoogleAttH}, all these models are in 32-dimensional vector space.

From the results, we have the following observations.
At first, the four hyperbolic-based models generally outperform their Euclidean variants 
and RotatE and TuckER, which are the state-of-the-art models for high-dimensional knowledge graph embeddings. 
This proves the effectiveness of hyperbolic models in low-dimensional knowledge graph embeddings.

RotL outperforms RotE and the other Euclidean-based models. Compared with RotE, the Hits@10 of RotL improves from 0.529 to 0.550 on WN18RR, and from 0.482 to 0.500 on FB15k237. 
Using a lightweight architecture, RotL even achieves similar prediction accuracy 
as RotH on FB15k237.
It indicates that the hyperbolic embeddings technology is
possible to be replaced with flexible addition and new distance function.
Using a novel two-layer transformation function, Rot2L 
further improves RotL and achieves the state-of-the-art results on two datasets.
Especially compared with RotH, MRR of Rot2L improves from 0.314 to 0.326 on FB15k237, and the Hits@1 increases from 0.428 to 0.434 on WN18RR. 

It should be noted that 
improving low-dimensional performance is much harder than that in the high-dimensional condition. Our experimental results prove the effectiveness of Rot2L, while the computational complexity of Rot2L is lower than RotH and AttH.

\begin{table*}[!tb]
\caption{Comparison of Hits@10 on relation-specific triples in WN18RR. $\Delta$(L-H) denotes the growth rate of RotL relative to RotH, and similarly $\Delta$(2L-H) denotes the growth rate of Rot2L. Higher $Khs_\mathcal{G}$ and lower $\xi_\mathcal{G}$ mean more hierarchical.}
\centering
\label{tab:4}
\resizebox{0.9\linewidth}{!}{
\begin{tabular}{lcc|cc|cc|cc}
\hline
\textbf{Relation}  & \textbf{$Khs_\mathcal{G}$} & \textbf{$\xi_\mathcal{G}$ } & \textbf{RotE} & \textbf{RotH} & \textbf{RotL} & \textbf{$\Delta$(L-H)} & \textbf{Rot2L} & \textbf{$\Delta$(2L-H)}\\
\hline
hypernym & 1.00 & -2.46 & 0.242 & 0.250 & 0.247 & \textcolor[rgb]{0,0.7,0}{-1.12\%} & \textbf{0.254} & {\color{red}1.44\%}\\
derivationally related form & 0.07 & -3.84 & 0.960 & \textbf{0.970} & 0.967 & \textcolor[rgb]{0,0.7,0}{-0.24\%} & 0.968 & \textcolor[rgb]{0,0.7,0}{-0.15\%}\\
instance hypernym  & 1.00 & -0.82 & \textbf{0.533} & 0.480 & 0.496 & {\color{red}3.42\%} & 0.467 & \textcolor[rgb]{0,0.7,0}{-2.57\%}\\
verb group   & 0.07 & -0.50 & \textbf{0.974} & \textbf{0.974} & \textbf{0.974}  & 0.00\% & \textbf{0.974} & 0.00\%\\
also see  & 0.36 & -2.09 & 0.634 & 0.643 & 0.625 & \textcolor[rgb]{0,0.7,0}{-2.78\%} &  \textbf{0.688} & {\color{red}6.95\%} \\
has part & 1.00 & -1.43 & 0.343 & \textbf{0.349} & 0.329 & \textcolor[rgb]{0,0.7,0}{-5.83\%} &0.334 & \textcolor[rgb]{0,0.7,0}{-4.17\%} \\
\textbf{member of domain usage} & 1.00 & -0.74   & 0.375 & 0.292 & \textbf{0.417} & {\color{red}42.86\%} & \textbf{0.417} & {\color{red}42.86\%} \\
\textbf{member of domain region} & 1.00 & -0.78  & 0.365 & 0.269 & 0.385 & {\color{red}42.86\%} & \textbf{0.442} & {\color{red}64.27\%}\\
similar to  & 0.07 & -1.00 & \textbf{1.000} & \textbf{1.000} & 0.667 & \textcolor[rgb]{0,0.7,0}{-33.34\%} & \textbf{1.000}& 0.00\%\\
member meronym  & 1.00 & -2.90 & 0.342 & 0.395 & 0.405 & {\color{red}2.49\%}& \textbf{0.451} & {\color{red}14.00\%} \\
synset domain topic of & 0.99 & -0.69 & 0.382 & 0.408 & 0.439 & {\color{red}7.53\%}& \textbf{0.443} & {\color{red}8.61\%}\\
\hline
\end{tabular}
}
\vspace{-2mm}
\end{table*}

\begin{table}[!tb]
\caption{The results of ablation experiments on (a) FB15k237 and (b) WN18RR.}
\centering
\small
\label{tab:3}
\begin{tabular}{lccc}
\hline
\multirow{2}*{\textbf{Methods}} & \multicolumn{3}{c}{\textbf{FB15K237}}\\
~ & \textbf{MRR} & \textbf{Hits@10} & \textbf{Hits@1}\\
\hline
Rot2L & \textbf{0.326}  & \textbf{0.503}  & \textbf{0.237} \\
Rot2Lw/oMid & 0.317 & 0.497 & 0.229\\
Rot2Lw/oDis & 0.316 & 0.492 & 0.226\\

RotL & 0.320  & 0.500  & 0.229 \\
RotLw/oDis & 0.316 & 0.495 & 0.227\\
RotE & 0.307 & 0.482 & 0.220\\
\hline
~&\quad\textbf{(a)}&~&~\\
\\
\hline
\multirow{2}*{\textbf{Methods}} & \multicolumn{3}{c}{\textbf{WN18RR}}\\
~ & \textbf{MRR} & \textbf{Hits@10} & \textbf{Hits@1}\\
\hline
Rot2L & \textbf{0.475}  & \textbf{0.554}  & \textbf{0.434} \\
Rot2Lw/oMid & 0.474 & 0.554 & 0.435 \\
Rot2Lw/oDis & 0.463 & 0.543 & 0.426 \\

RotL & 0.469  & 0.550  & 0.426 \\
RotLw/oDis & 0.467 & 0.549 & 0.420 \\
RotE & 0.463 & 0.529 & 0.426\\
\hline
~&\quad\textbf{(b)}&~&~
\end{tabular}
\vspace{-6mm}
\end{table}

\subsection{Ablation Studies}
We further 
conduct 
a series of ablation experiments to evaluate the different modules 
of 
our models.
Two main improvements should be evaluated: 1) the new distance function (Dis) in Eq. \ref{eq3.1} and 2) the middle activate function (Mid) in Eq. \ref{eq5}.
Accordingly, 
we test 
the variants by eliminating one of the two functions (e.g., Rot2Lw/oDis by removing the distance function).
The other parts, such as flexible addition and stacked transformations, can be verified by comparing RotE 
(a Euclidean-based variant of RotH), RotL and Rot2L.
The experimental results are shown in Table \ref{tab:3}.

From the results, we can see that
Hits@10 of Rot2L are higher than Rot2Lw/oDis on the both datasets, which proves the effectiveness of the distance function. 
Similar result is also shown in RotL, but the improvement is relatively small.
Hits@10 of Rot2Lw/oMid are lower than that of Rot2L on FB15k237, while having no obvious difference on WN18RR.
This indicates that the activate function is more effective on FB15k237, which contains much more relations than WN18RR.
For FB15k237, RotL outperforms Rot2Lw/oMid, indicating that facing complex relationships, the pure two-layer transformation is no better than a single layer. 
This further validates the contribution of the activate function. 
Comparing RotLw/oDis and RotE, the impact of flexible addition is obvious. Using a simple scaling operation, the flexible addition provides a 1\% and 2\% improvements of Hits@10 on the two datasets, respectively.

Overall, the experimental results indicate the effectiveness of the major modules of our proposed models in this paper. 
Based on the same Euclidean space, our 
RotL and Rot2L models have a significant performance improvement compared to the RotE model.

\subsection{Efficiency Analysis}
\label{sec:3.3}
We analyze and compare the computational complexity among RotE, RotH, RotL, and Rot2L in this section.
In terms of time complexity, as shown in Fig. \ref{fig:1}, RotL is much faster than RotH mainly because that the Flexible Addition requires only a quarter of the computational cost of the M{\"o}bius addition. Although Rot2L repeats the rotation-translation transformation twice, its computational cost is still lower than that of RotH. 

In terms of space complexity, a slight difference is shown in the number of relation parameters. Because the parameters related to entities, including entity embedding vectors and entity biases, are the same in the four models and occupy the vast majority of total parameters, RotH requires the most relation parameters, $(3N_r+1)d$, including three relation transformation vectors and the learnable curvature for different relations. By contrast, RotE and RotL cost smaller, which are $2N_rd$ and $2(N_r+1)d$, and the extra part of RotL comes from $\alpha$ in Flexible Addition. Although using an effective parameterization strategy, Rot2L still requires two shared vectors and another $\alpha$-related vector.
Its relation parameter amount is $(2N_r+5)d$. As the relation number $N_r$ is always greater than four, the Rot2L model 
requires fewer parameters than RotH.

In summary, 
the RotL and Rot2L models are 
highly efficient and better than the RotH model in both time complexity and space complexity.

\section{Discussion}
\label{sec:5}
In this section, we further discuss several important questions on the RotL and Rot2L models.

\vspace{2mm}
\noindent\textbf{Q1: Which parts of predictions are improved in our models comparing with RotH?}

We measure the link prediction performance of relation-specific triples on WN18RR, shown in Table \ref{tab:4}, to analyze the improvements of the Rot2L model. The results are generated in the 32-dimensional condition.
Comparing in different relations, RotE and RotH have their own strengths.
RotH has better Hits@10 in most 
relations but is weaker than RotE in the ``member of domain usage'' and ``member of domain region'' relations.
RotL performs well like RotH, but fails to predict the ``similar to'' relation. 
As the optimal model, Rot2L obtains the best Hits@10 in 8 out of 11 relations and only 
has a small decrease in the other three relations.
It should be noted that RotL and Rot2L effectively improve on the two ``member of'' relations, comparing to RotH. Especially, Rot2L achieves 42.86\% and 64.27\% improvements than RotH on these relations. 
Achieving the flexible normalization of RotH in the Euclidean space,
our models perform well in both RotH-dominant and RotE-dominant 
relations.

\vspace{2mm}
\noindent\textbf{Q2: Can our models encode hierarchical patterns like hyperbolic-based models?}

As RotH has been proved on the benefits of hyperbolic geometric on hierarchical relations,
we further analyze whether our models can still preserve this property.
Following the work of Chami et al. \cite{GoogleAttH}, we utilize the Krackhardt hierarchy score ($Khs_\mathcal{G}$) and estimated curvature ($\xi_\mathcal{G}$) as metrics. The related results can be found in Table 4, in which a relation with higher $Khs_\mathcal{G}$ and lower $\xi_\mathcal{G}$ is more hierarchical.

In terms of non-hierarchical relations, such as ``verb group'', Euclidean-based and hyperbolic-based RotH have similar performances.
In terms of hierarchical relations satisfying $Khs_\mathcal{G}=1$, we observe that hyperbolic embeddings work better on relations having low $\xi_\mathcal{G}$, such as ``hypernym'', ``has part'', and ``member meronym''. Meanwhile, RoE and RotL outperform RotH in relations having relative higher $\xi_\mathcal{G}$, such as ``instance hypernym'', ``member of domain usage'', and ``member of domain region''.
Compared with the other three models, Rot2L obtains the best Hits@10 in most relations and works effectively on hierarchical relations with different $\xi_\mathcal{G}$.

The results indicate that the simplified models, RotL and Rot2L, have a good ability to encode hierarchical relations. They preserve the good properties of both hyperbolic geometric and the Euclidean-based RotE.

\begin{figure}[!bt]
\centering
\includegraphics[width=0.47\textwidth]{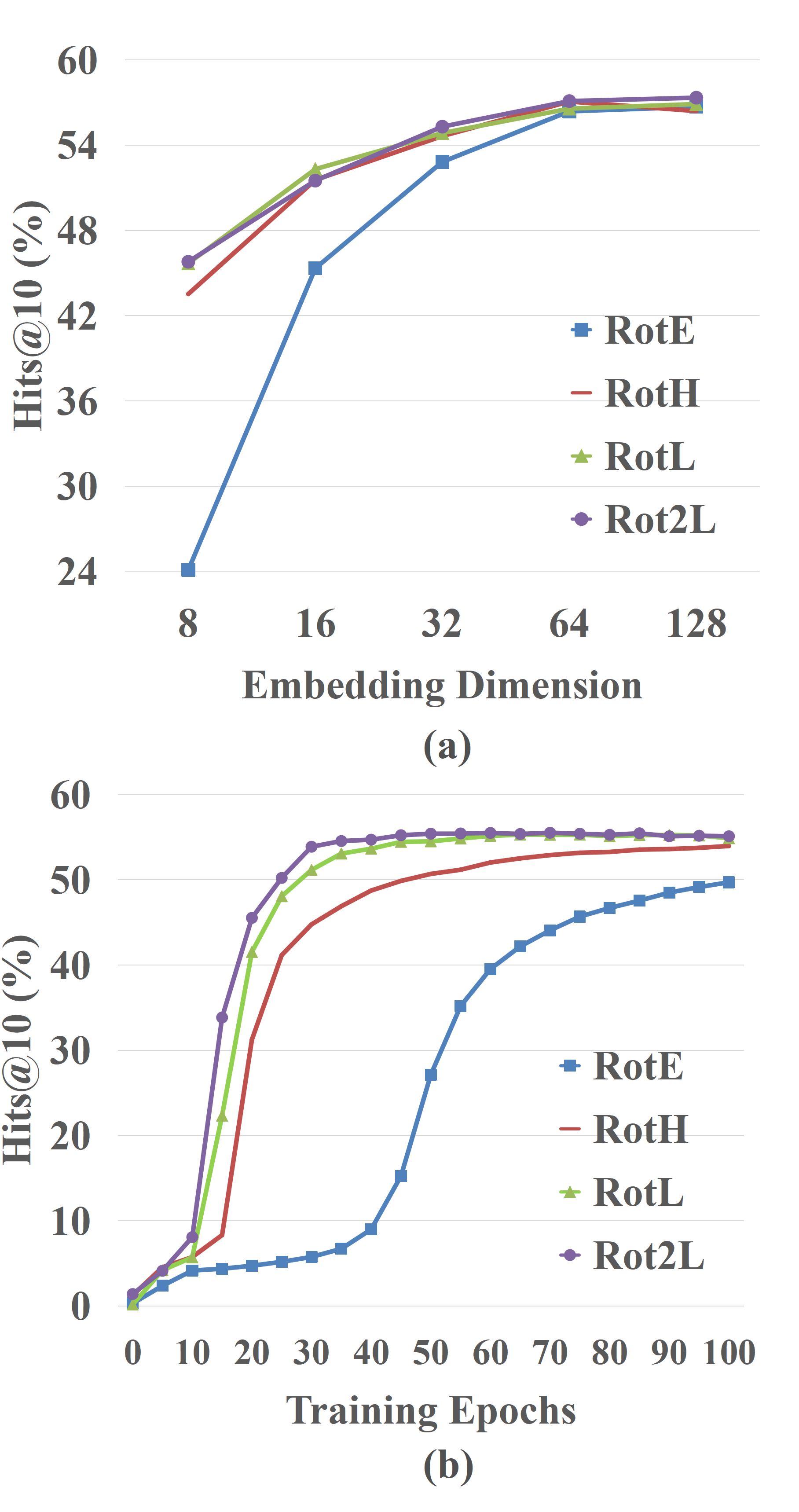}
\caption{(a) The Hits@10 of four models on WN18RR with different embedding dimensions. (b) The changes of the Hits@10 on WN18RR as training proceeds.}
\label{fig:3}
\vspace{-4mm}
\end{figure}

\vspace{2mm}
\noindent\textbf{Q3: How about the model performance in other embedding dimensions?}

We further compare the four models in different dimensions from 8 to 128.
The experimental results are shown in Fig. \ref{fig:3}(a).
The prediction accuracy of the four models improves with 
the growth of the embedding dimensions. 
When the dimensions are lower than 32, RotE is obviously weaker than the others, 
but it performs well in the high-dimensional condition. 
Except for RotE, the other three models obtain similar results under high dimensions, but there are still some differences. 
Specifically, 
RotL performs better in lower dimensions and 
achieves the best Hits@10 under 16 dimensions. The advantage of Rot2L is shown with the increase of dimensions: Rot2L outperforms the other two after 32 dimensions.
The experimental results prove that the stacked transformations in Rot2L have stronger representation capacity in the high-dimensional condition.

\vspace{2mm}
\noindent\textbf{Q4: Can our models accelerate the training speed?}

Fig. \ref{fig:3} (b) shows the convergence of the training process for the four models under 32 dimensions. We can observe that RotE increases slowly in the first 40 epochs and converges later than the others. 
RotH converges faster than RotE, which is previously regarded as the contribution of hyperbolic space.
From our experimental results, 
it is clear that both RotL and Rot2L show similar performance. 
RotL, which has little difference from RotE in structure, shows a much faster training speed.
Although RotE takes less time over one epoch, RotL can achieve higher performance with less training epochs. 
Comparing RotH and Rot2L, we can find that Rot2L precedes in almost every epoch. In the 25th epoch, Rot2L already achieves similar performance 
as the final performance of RotH.
The results indicate that our models can replace the hyperbolic RotH model 
with comparable prediction accuracy and training speed.

\section{Related Work}
\label{sec:6}
\subsection{Knowledge Graph Embeddings}
Various KGE models have been proposed using different scoring functions, such as translation-based TransE \cite{TransE}, factorization-based ComplEx \cite{Complex} and CNN-based ConvE \cite{ConvE}.
With the rise of deep learning, several DL-based methods have been proposed, such as ConvKB \cite{ConvKB} and CompGCN \cite{CompGCN}.
Balazevic et al. \cite{TuckER} propose a linear model based on Tucker decomposition of the binary tensor representation of knowledge graph triples.
RotatE \cite{RotatE}, inspired by Euler's identity, represents a relation as the rotation operation between the head and tail entities. 
DihEdral \cite{DihEdral} introduces rotation and reflection operations in dihedral symmetry group to construct the relation embeddings. 
Similar to the previous
approaches, 
these models utilize high-dimensional embedding vectors while designing a new score function to better distinguish the 
triples.

\subsection{Hyperbolic Embeddings}
Hyperbolic geometry has recently drawn wide attention because of its potential to learn parsimonious representations of symbolic data by simultaneously capturing hierarchy and similarity \cite{HBEmbed0,HBEmbed1,HBEmbed2}.

Recently, some researchers start to apply hyperbolic embedding in the KGE domain.
Bala\v{z}evi\'{c} et al. \cite{MuRP} propose the MuRP model to embed KG triples in the Poincar\'{e} ball model of hyperbolic space using the M\"{o}bius matrix-vector multiplication and M{\"o}bius addition operations.
Similarly, Kolyvakis et al. \cite{HyperKG} extend the translational models by learning embeddings of KG entities and relations in the hyperbolic Poincar\'{e}-ball model.
Sun et al. \cite{EMNLP20HBKGE} propose a hyperbolic relational graph neural network to capture knowledge associations for the KG entity alignment task.
Chami et al. \cite{GoogleAttH} employ rotation and reflection operations to replace the multiplication operation between the head entity and relation vectors, and propose a series of hyperbolic KGE models with trainable curvature, including RotH, RefH, and AttH.

Comparing with the existing hyperbolic KGE models, our model simplifies the hyperbolic calculations to improve computational efficiency while achieving competitive performance.

\section{Conclusion}
\label{sec:7}
The recently proposed hyperbolic-based models 
achieve great prediction accuracy
in low-dimensional knowledge graph embeddings, but require complicated calculations for hyperbolic embeddings. 
In this paper, we analyze the effective components in those models and propose a lightweight variant based on Euclidean calculations. After simplifying the M{\"o}bius operations in RotH, our proposed RotL model achieves a competitive performance, which 
saves half of the training time. Using a two-layer stacked transformation, we further propose Rot2L that outperforms the state-of-the-art RotH model in both prediction accuracy and training speed.

These positive results encourage us to explore 
further research activities in the future.
We will theoretically analyze the effectiveness of flexible normalization 
in the low-dimensional KGE tasks.
For the stacked transformations in Rot2L, we will explore multiple-layer architectures and evaluate more different transformation forms. Finally, we plan to apply our models on real-world knowledge graphs in different domains such as mobile healthcare, smart cities, and mobile e-Commerce.

\section*{Acknowledgments}
This research is supported by the National Natural Science Foundation in China
(Grant: 61672128) and the Fundamental Research Fund for Central University (Grant: DUT20TD107).
Quan Z. Sheng has been partially supported by Australian Research Council (ARC) Future Fellowship Grant FT140101247, and Discovery Project Grant DP200102298.

\bibliography{output} 
\bibliographystyle{acl_natbib}

\end{document}